\def\year{2020}\relax
\documentclass[letterpaper]{article} 
\usepackage{aaai20}  
\usepackage{times}  
\usepackage{helvet} 
\usepackage{courier}  
\usepackage[hyphens]{url}  
\usepackage{graphicx} 
\usepackage{graphicx}  
\usepackage{amsmath,amssymb}
\usepackage{setspace,multicol,multirow,enumerate,textcomp}
\usepackage[dvipsnames]{xcolor}

\title{Experimental Studies in General Game Playing: An Experience Report}
\author{Jakub Kowalski, Marek Szyku{\l}a\\  
University of Wroc{\l}aw, Poland \\
jko@cs.uni.wroc.pl,  msz@cs.uni.wroc.pl 
}
 
\begin{document}

\maketitle

\begin{abstract}
We describe nearly fifteen years of General Game Playing experimental research history in the context of reproducibility and fairness of comparisons between various GGP agents and systems designed to play games described by different formalisms. We think our survey may provide an interesting perspective of how chaotic methods were allowed when nothing better was possible. Finally, from our experience-based view, we would like to propose a few recommendations of how such specific heterogeneous branch of research should be handled appropriately in the future. The goal of this note is to point out common difficulties and problems in the experimental research in the area. We hope that our recommendations will help in avoiding them in future works and allow more fair and reproducible comparisons.
\end{abstract}

\section{Introduction}

As an alternative for the research trying to solve particular human-created games, like chess \cite{Campbell2002Deep}, checkers \cite{Schaeffer2007Checkers}, or go \cite{Silver2016Mastering}, General Game Playing (GGP) domain has been established to learn computers to play any given game, in particular, the one with previously unknown rules. 
Although the main idea can be traced back to the famous General Problem Solver from 1959 \cite{Newell1959Report}, the area really started developing with the call for the Stanford's GGP competition in 2005 \cite{Genesereth2005General}.

Since that time, more GGP systems and formal game description languages emerged, and for all of those systems, many agent's descriptions, algorithms, and theoretical analyses have been published. However, the domain fragmentation and complexity of used solutions cause the recurring issue of the once-done-never-rerun experiments. Thus, they were reported, and in most cases, could not be reproduced or even verified in any convincing way.

In this paper, we are presenting an excerpt from experimental studies from nearly fifteen years of general game playing history in the context of research reproducibility \cite{gundersen2018reproducible}. We are investigating the fairness of comparisons between various GGP agents and also systems designed to play games described by different languages, providing a large number of descriptive examples. 
In particular, we demonstrate a very recent collection of works that visualizes how important it is to ensure that the results are well described and verifiable -- which in the other case may even cause a debate levering the correctness of the paper.

From a broad perspective, our goal is to provide a perspective of chaotic methods that were allowed when nothing better was possible. Finally, from our experience-based view, we would like to propose a few recommendations of how such specific heterogeneous branch of research should be handled appropriately in the future, and how to avoid at least some of the issues indicated in this report.

\section{General Game Playing Domain}

The oldest General Game Playing (GGP) approaches focused on generalizations of chess-like games, which can be consistently described under one formalism, and then played or even procedurally generated \cite{Pitrat1968Realization,Pell1992METAGAME}.

The real advancement, and formal establishing of GGP as a proper research domain, was due to the announcement of Stanford's Game Description Language (GDL) and associated International General Game Playing Competition (IGGPC) in 2005 \cite{Genesereth2005General,Love2008General}.

GDL can describe any turn-based, finite, and deterministic $n$-player game with perfect information. It is a high-level, strictly declarative logic language, based on Datalog \cite{Abiteboul1995Foundations}. The language does not provide any predefined functions, so every predicate encoding the game structure like a board or a card deck, or even arithmetic operators, must be defined explicitly from scratch.
Of course, this should not be seen as a drawback of GDL, because due to that, it describes games in a very knowledge-free way, stimulating the development of knowledge-inference methods from a general game description.

For quite a long time, the entire GGP domain was equated to Stanford's GGP \cite{Genesereth2014General}. This is definitely the most influential system, resulting in many valuable research and algorithm advancements (e.g., for MCTS \cite{Browne2012ASurvey}). Multiple GDL extensions have been designed, e.g., GDL-II introducing randomness and imperfect information \cite{Thielscher2010AGeneral}, rtGDL removing from the system the turn-based restriction \cite{Kowalski2016TowardsARealtime}, or GDL-III for describing games with imperfect information and introspection \cite{thielscherIJCAI17}.

After Stanford's GGP gained popularity, more general game playing formalisms had been designed. 
Some of them are aiming to be as general as GDL but faster or more concise.
Some are describing significantly simpler classes of games, but because of that, giving more information to the agents.
Finally, some are entirely unrelated, aiming to represent, e.g., real-time video games.

\subsection{A variety of GGP languages}

In this section, we will briefly introduce other GGP formalisms that will be mentioned throughout the paper.

TOSS \cite{Kaiser2011FirstOrder}, proposed as a GDL alternative, is based on first-order logic with counting. The language structure allows more accessible analysis of games, as it is possible to generate heuristics from existential goal formulas.

Simplified Boardgames \cite{Bjornsson2012Learning} describes chess-like games using regular expressions to encode movement of pieces. Its expressiveness is very limited, but the language is concise, and conceptually well defined. Thus it is relatively simple to use in some advanced tasks like a procedural generation of game rules \cite{Kowalski2016EvolvingChesslike}.

Ludi system was designed for the sake of the procedural generation of games from a restricted domain of combinatorial games \cite{Browne2010Evolutionary}. Its successor Ludii \cite{piette2019ludii} (currently under development), aims to describe any traditional strategy game throughout recorded human history. The system is planned to be used to chart the historical development of games and also explore their role in the development of human culture.

Regular Boardgames (RBG) \cite{Kowalski2019RegularBoardgames} is a novel GGP system based on the theory of regular languages. 
Like GDL, it provides only a few generic mechanisms that already allows describing the class of all finite deterministic turn-based games with perfect information.
But instead of logic, it uses a transition system of modifying the game state that naturally occurs in games.
Therefore, it allows concise encoding and effective playing games with complex rules and with large branching factor such as arimaa, go, and international checkers -- which is impossible in, e.g., Stanford's GDL.

On the contrast of the beforementioned formalisms, General Video Game AI (GVGAI) domain is strictly focused on representing and playing real-time Atari-like video games \cite{perez2019general,gvgaibook2019}. It recently gained much popularity due to its (relative) simplicity, so the organizers run a few multitrack competitions per year. Instead of game descriptions, as in the case of Stanford's GGP, the GVGAI competition framework provides a forward model as a programming structure. Thus, the agents do not need to (and cannot) parse and understand game semantic directly from its rules.

\section{Comparing Agents Efficiency\\ in Stanford's GGP}

To visualize the Stanford's GGP domain difficulties and progress in the quality of presenting results, we discuss the development of the branch devoted to improving the reasoning efficiency of GDL.
In contrast with the theoretical research, which usually does not require experiments or they are performed to visualize the effect of the proposed solution unrelated to other solutions, this branch should always provide reliable comparisons between existing approaches.

The growing number of competition-ready GGP agents and problems with slow standard Prolog-based GDL reasoning resulted in many proposals to speed-up the logic resolution and compute game states faster. Soon, this became a crucial aspect of both research progress and success during the tournaments. Because GDL is a subset of first-order-logic Datalog language with specific keywords and information flow added, the possibilities are vast; just to mention compilations to other languages, partial computation of possible fact occurrences (\emph{instantiation}), optimizing data structures for fast querying, or even putting GDL reasoners into a hardware \cite{Siwek2018ImplementingPropositional}.

\subsection{Context and Limitations}

For the sake of completeness, we would like to point out reasons that shape Stanford's GDL research as it was.

Probably, the most important issue was full freedom in available technologies. The specified http-based communication protocol allows using any programming language (or a mix of programming languages), any third-party resources, and any hardware -- which caused situations like a game between agents running on a single laptop vs.\ on a cluster of computers.

The requirement that the game manager communicates with the player via the TCP/IP connection causes another set of issues. For example, all players had to take into account the communication lag. Thus, all messages were usually sent a few seconds before the timelimit. Given that for some games these limits oscillates in 15-30 seconds, these few seconds may have some non-negligible impact on the player's performance, yielding a disadvantage depending on the geographical localization.
Also, the communication architecture forces player agents to be servers (while the game manager was a client-type application), which requires public IP to play against other online agents.

Lastly, research on Stanford's GGP started in 2005 and grew rapidly until about 2015--2017. During this time, the necessity for fully reproducible research, although needed, was not as clearly defined as today. Also, managing ongoing open-source projects was slightly more complicated. The most widely used open-source GGP Base project containing i.a. example players, games repository, rules validator, game manager allowing agent vs.\ agent and agent vs.\ human games, was started in late 2013 \cite{Schreiber2013GGPBase}.

\subsection{Overview of the Research}

Let us then briefly present and analyze several works concerning the efficiency improvements from the point of view of experiments and their supposed reproducibility.

\textbullet\ \cite{Waugh2009Faster} 
is the first published GDL compiler. The author translates the GDL descriptions into a compilable C++ code. Experiments compare the efficiency of the proposed solution against example YAP Prolog-based GDL interpreter on four games: tic-tac-toe, chess, checkers, and connect4. They provided a number of examined states and performed simulations over 5 seconds using a flat MC approach. Additionally, the author tested the version with transposition tables and differentiate results for game stages described as early, middle, and end. Compilation times were not provided.

\textbullet\ \cite{Kissman2010Instantiating} proposes a process of instantiation, i.e., grounding all occurrences of variables within the GDL description. Instantiated rules, although syntactically larger, are usually much faster to process during the gameplay. (The instantiation algorithms become a subbranch of GDL research, c.f.\ e.g., \cite{Vittaut2014Efficient}.)

The experiments were performed to test how many of the 171 games from the Dresden GGP server were possible to ground via a Prolog-based approach, and via their dependency graph technique. Also, for 10 games, the performance of SWI-Prolog reasoner on standard and instantiated game rules was compared using a number of Monte Carlo simulations as a measure.

\textbullet\ \cite{Saffidine2011AForward} presents a forward chaining GDL compiler into Ocaml. The results compare a number of Flat Monte Carlo playouts performed within 30 seconds against the YAP Prolog interpreter. They contain compilation times and the initial and final file sizes. The test set consists of 9 games, including breakthrough, connect4, and three variants of tictactoe.

\textbullet\ \cite{Kaiser2011FirstOrder}\footnote{\url{http://toss.sourceforge.net/}} presents a method of rewriting GDL into previously mentioned Toss system, which makes use of CNF and DNF conversions, SAT solvers, and first-order logic model checkers.

The authors compare the efficiency of their system by playing against Fluxplayer \cite{Schiffel2007Fluxplayer}, and showing winrates on four games: breakthrough, connect4, connect5, and pawn whopping. The Toss search algorithm was based on constant-depth alpha-beta.

\textbullet\ \cite{Kowalski2013GameDescription} presents a highly optimized bottom-up GDL to C++ compiler with multiple features, including predicates flattening and optimized data structures to fasten query times.
The results include compilation times, the number of simulations, and the number of computed game states per second during Flat MC. The comparison was made against ECLiPSe Prolog reasoner, and included 6 games: tictactoe, blocker, connect4, breakthrough, checkers, and skirmish.

\textbullet\ \cite{Schofield2013High} is another forward chaining GDL compiler, which is, in some sense, an improved version of the previous work. It contains a more detailed experiment section, measuring compilation times, the number of states per second, and the number of playouts per 30 seconds of Flat MC playouts on 19 games. The results were not directly compared to any baseline.

\textbullet\ \cite{Bjornsson2013Comparison} attempts to be a complex comparison on existing GDL reasoners. Actually, it compares Fluxplayer \cite{Schiffel2007Fluxplayer} and CadiaPlayer \cite{Bjornsson2009CadiaPlayer}\footnote{\url{http://cadia.ru.is/wiki/public:cadiaplayer:main#cadiaplayer_source}}, presenting one test against some basic reasoners (also including GGP Base package \cite{Schreiber2013GGPBase}), but omits head-on comparison with any of the previously mentioned published approach.

The results for Fluxplayer and CadiaPlayer include visited nodes per second for two search algorithms: Monte Carlo and Min-Max. There are also tests versus game-specific reasoners. The comparison with other engines does not include numbers, only a bar chart with relative speeds. All experiments were performed on a set of 12 testgames.

\textbullet\ \cite{Swiechowski2014Fast} describes various optimizations of game encoding, including rewriting queries and specific memory representation. The experiments were performed on 28 games, and compared a number of random Monte Carlo simulations versus ECLiPSE Prolog and YAP Prolog.

\textbullet\ \cite{Sironi2016Optimizing} introduces optimizations on propositional networks, which are a very efficient alternative representation of a GDL code. Experiments were performed on 13 games, testing multiple variants of the algorithm.
The results include the number of nodes per second in flat MC, initialization times, and the size of the resulting network. The work also compares the speed and the win-percentage versus GGP-Base Prover \cite{Schreiber2013GGPBase}.

\subsection{Findings}

The most conspicuous issue is the almost complete lack of direct comparisons. In our opinion, this is due to the following reasons. First, most of the systems were not available as open-source (partially because of the domain's competitiveness, partially for other reasons like reluctance to share messy code). However, some of them, e.g., CadiaPlayer or Toss, are available for quite a long time, and they are still not used as a testbase.

This is because of the second reason -- GGP agents are very complex systems, and using off-the-shelf code causes many problems. They are usually not designed to be operated by anyone except the authors, and sometimes they are even prepared to work on a specific hardware / network architecture. Thus, even with the source, modifying them to ensure fair comparison (e.g., the same search algorithm used) is a very tiresome work. 
A remedy for these issues would be publishing a full list of project dependencies or a Dockerfile, allowing its fully accurate setup. 

On the positive sides, most of the enumerated experiments use a similar comparison approach, i.e., the number of playouts or the number of visited states during Flat Monte Carlo simulations. Also, usually, the papers mention the hardware specifications, of course, with various levels of details.
Often, there is missing information about the amount of RAM and operating system, not to mention nearly non-existing detailed, but still important, data about software versions and the number of running threads.

However, the flat MC algorithm is commonly underspecified.
While it is generally understood, it can vary in some subtle details, e.g., whether the final goals of the players are computed after each random playout.

We can observe a common trend for sources of the game descriptions: for the earlier works, it is Dresden GGP Server \cite{DresdenGPPServer}; and for the later ones, GGP Base repository \cite{Schreiber2016Games}. Providing the location of the source should allow to accurately know the GDL codes used, which is very important as in GDL even small encoding differences may influence the reasoning difficulty. However, as for some games multiple versions of the rules exist in the same repository, the game name should be stated in a form allowing unambiguous matching.

Let us observe that the sets of testgames used in experiments are not standardized in any way. Firstly, they are usually minimal, especially in early works, due to the time required for gathering enough reliable samples. Secondly, the choices of test games are often discordant. Some games are most common than others, but finding a good testset to compare results with other approaches remains a difficult task. Just to mention that the only game commonly tested in all the above works is tictactoe.

The last observation is that with time, the quality of the presented experiments improves.  More recent publications tend to be better documented and contain more insightful experiments. However, still many solutions are not released as open-source, so reproduction of the results is not possible.

\subsection{Comparison to GVGAI}

As an example of a newer and better-organized GGP competition, we will briefly describe General Video Game AI, launched in 2014, and running multiple-track competitions since then \cite{perez2019general}.
With the downside of forcing agents to be written in Java or Python only, GVGAI standardizes the agent's structure and communication protocol more firmly. There is no net-based communication. To take part in the competition, the agent's properly formatted sources have to be sent to the competition page, where they are automatically run and evaluated.

The framework itself is open-sourced and regularly updated. Thus, one official game repository is available, every game is labeled, and its rules are corrected if necessary. Players' agents are not necessarily open, however some data about their performance, e.g., score on each of the global sets of testing games, is publicly available on the website.

Thus, although still far from the ideal, the example of GVGAI shows that it is possible to organize a general game playing competition that will allow easier data exchange, more insights into the results obtained by the agents, and thus better reproducibility.


\section{Comparing Different GGP Formalisms}

The existence of multiple GGP languages arises a natural set of questions: which language is better -- more universal, more efficient, more readable, easier for learning or PCG; do agents in one system have some natural advantage over the agents from the other system; can we automatically translate rules between these languages, etc.?
Because obtaining any reliable answer to these questions is very difficult and requires much work, both conceptually and programmatically, there is relatively little research tackling these problems.

\subsection{Syntactical translations}

Automatic translation of game rules from one GGP language to the other, of course, depends on the generality of those languages. So far, apart from the theoretical mapping of formalisms like in \cite{Thielscher2011Translating}, only one-sided translations from simple GGP languages to versions of Stanford's GDL were proposed.
All works described here present experiments in which translated rules are fully playable games (in either GDL or Toss).

\textbullet\ First published translation concerns mapping from GDL to Toss \cite{Kaiser2011Translating}. The authors provide a detailed description of the translation, with multiple informative examples for each step, and formally prove its correctness. The obtained Toss rules are described as ,,inefficient'' and ,,verbose'', compared to the ones created manually, but no numbers are provided. A brief experimental section contains the results on matches between Toss players using manual and translated rules on two games only: one ended with 100\% of ties, the other one with 10\% advantage for a manually written description.

\textbullet\ In~\cite{Kowalski2014EmbeddingACard}, the author describes a mapping from very domain-specific Card Game Description Language \cite{Font2013ACard} into GDL-II (GDL extension with randomness and imperfect information). The work puts an overview of the translation rules, proves its correctness, and measures dependence between the complexity of both descriptions.
Experiments are performed on three game codes (plus some variants) that were originally published for the card language. The main conclusion was aimed at the verbosity of GDL, as using the proposed translation, even a very simple game, requires hundreds of rules to be encoded. The system is not open-sourced, nor any more detailed analysis of the translation is available.

\textbullet\ On the other hand, \cite{Sutowicz2016Simplified} presents a translation from Simplified Boardgames \cite{Bjornsson2012Learning} into standard GDL, which aims at producing code that is computationally optimized. In particular, when translating regular expressions into GDL rules, it tries to share partial expressions to reduce code redundancy. The thesis contains proofs of bounds: on the size of resultant code and on the time complexity of the algorithm. The experiments show the improvement factors of applied optimizations. Tests have been performed on 11 chess-like games, including procedurally-generated ones and specially handcrafted examples. What is worth mentioning, although without documentation, the translator source is available online\footnote{\url{https://github.com/uicus/sbg2gdl}}.

\subsection{Game playing comparison}

Yet another, even more challenging and definitely more error-prone task is to experimentally assess language properties via the agent's performance when both agents belong to different GGP systems.

\textbullet\ In previously mentioned Toss publication \cite{Kaiser2011FirstOrder}, the authors, to show the benefits of their formalism, defined several board games in it and created move translation scripts allowing playing against GGP agents. Thus, the comparison works under the assumption that each game is encoded optimally for both languages. (Which gives some advantage for Toss, where it is easier to derive game heuristics from the game rules automatically.)

Experiments were performed between two different search algorithms, Toss and Fluxplayer, run on different and unknown hardware on four games -- with one clear win for Toss, one for Fluxplayer, and two complete ties. So, apart from that, the presented solution is ,,good enough to win against a state of the art GGP player'', any more insightful conclusions are impossible to make.

\textbullet\ A very similar approach, aiming to assess the progress of Stanford's GGP by comparing state-of-the-art players with an exemplary agent from much simpler Simplified Boardgames class, has been taken in~\cite{Kowalski2015TestingGeneral}.
Here, the authors developed a bridge, allowing simplified boardgames agent to play using Stanford's GGP competition protocol.

A simple min-max player with evaluation function obtained by temporal difference learning was paired against two GGP players, including 2014 IGGPC champion Sancho, on four chess-like games -- however, all three agents were launched on different hardware. The interesting aspect of the experiment was that in Simplified Boardgames the agent is aware of exiting pieces and board, thus it can reason about their values. In Stanford's GGP, those concepts are not so clearly visible, so in multiple tests, the simple player easily won against these advanced GDL-based agents.

\textbullet\ Regular Boardgames is yet another GGP language that proves to be faster than GDL \cite{Kowalski2019RegularBoardgames}. To visualize the advancement, the authors matched the rules of 12 games in both languages and performed experiments using both perft (computing the whole game tree to a fixed depth) and Flat MC counting the number of visited states. Two implementations of GDL engine were used: ECLiPSe Prolog (withing the gamechecker tool), and propnets from \cite{Sironi2016Optimizing}. The system is available as open-source with all the game codes used for the experiment\footnote{\url{https://github.com/marekesz/rbg/}}.

\subsection{Case study: comparing the efficiency of Ludii and Regular Boardgames}

The last example that we describe is a recent comparison \cite{piette2019empirical,piette2019ludii} of three different GGP languages (in particular, the efficiency of reasoning): Ludii, Regular Boardgames, and GDL.
The series of works is recent and leads to some interesting conclusions regarding conducting and reproducing experimental research.
We performed a detailed analysis of the experiments, trying to reproduce them.

It turned out that they are a good visualization of possible issues when comparing general game playing systems and also of difficulties coming out during a reproduction attempt.
We highlight some difficulties in methodologies and carried experiments that can significantly distort the results and actually turn the conclusions into the opposite.
This example is yet another evidence of how important it is, for any sort of experimental justification, allowing its easy and undoubtful repetitiveness.
We present this analysis in the hope of avoiding similar problems in any further research of this kind, allowing comparisons as fair as possible, which do not cause the need for questioning the results.

\subsubsection{Reproduction attempt}

Because of the unavailability of the Ludii version used in~\cite{piette2019empirical}, we performed an analysis based on one of the publicly released (and newer) Ludii version.

When we took Ludii games with the corresponding names according to the benchmark and analyzed their rules, we found out that only 5 out of 14 games have fully equivalent rules to those existing in RBG 1.0.
In most of the remaining cases, when the rules embed a different game variant than that existing in RBG 1.0, we made an attempt to reimplement that variant with more corresponding rules to the version in Ludii (but still only to some practicable extent).
Then we performed our benchmark.
A comparison of the results from both experiments is shown in Table~\ref{tab:experiments-mc}.
For the first two games, we visualize the problems caused by a mismatch between games that differ only by a variation of rules.
Our detailed technical analysis of this study is available at~\cite{Kowalski2019ANote}.

\begin{table*}[!ht]\renewcommand{\arraystretch}{1.2}\small
\newcommand{\rowt}[1]{\multirow{2}{*}{#1}}
\newcommand{\rowtt}[1]{\multirow{3}{*}{#1}}
\newcommand{\col}[1]{\multicolumn{1}{c|}{#1}}
\newcommand{\colL}[1]{\multicolumn{1}{c||}{#1}}
\newcommand{\colc}[1]{\multicolumn{1}{|c|}{#1}}
\newcommand{\colcL}[1]{\multicolumn{1}{|c||}{#1}}
\newcommand{\colt}[1]{\multicolumn{2}{c|}{#1}}
\newcommand{\coltL}[1]{\multicolumn{2}{c||}{#1}}
\newcommand{\coltt}[1]{\multicolumn{3}{c|}{#1}}
\newcommand{\colttt}[1]{\multicolumn{4}{c|}{#1}}
\renewcommand{\r}[1]{\textcolor{red!60!black}{#1}}
\renewcommand{\l}[1]{\textcolor{blue}{#1}}
\newcommand{\fa}{*}
\newcommand{\nda}{\hphantom{\dag}}
\newcommand{\ns}{\hphantom{*}}
\caption{The original and reproduced results of the efficiency of reasoning in RBG, Ludii, and GDL for the \textbf{flat Monte Carlo} test.
The values are the \textbf{numbers of playouts per second}.}\label{tab:experiments-mc}
\begin{center}\begin{tabular}{|l|r|r|r||l|r|r|r|}\hline
\multicolumn{4}{|c||}{Results from \cite{piette2019empirical}}&\multicolumn{4}{c|}{Results from \cite{Kowalski2019ANote}}\\\hline
{\bf Game}      & {\bf RBG 1.0}&     {\bf Ludii}&        {\bf GDL}&             {\bf Game}&{\bf RBG 1.0/1.0.1}&{\bf Ludii 0.3.0}&{\bf GDL propnet}\\\hline
\rowt{Amazons}  &    \rowt{625}&\rowt{\l{4,349}}&       \rowt{185}& Amazons-orthodox      &             569\ns&   \emph{n/a}\nda&               4 \\\cline{5-8}
                &              &                &                 & Amazons-split         &       \r{8,798}\ns&        3,859\nda&             365 \\\hline 
\rowt{Arimaa}   &   \rowt{0.11}&  \rowt{\l{714}}&\rowt{\emph{n/a}}& Arimaa-orthodox       &            0.14\ns&   \emph{n/a}\nda&      \emph{n/a} \\\cline{5-8}
                &              &                &                 & Arimaa-split          &         \r{666}*  &          446\dag&      \emph{n/a} \\\hline
Breakthrough    &    \r{16,694}&           4,741&            1,123& Breakthrough          &      \r{19,916}\ns&        3,546\nda&           2,735 \\\hline
Chess           &           714&         \l{720}&             0.06& Chess-fifty move      &         \r{523}*  &           14\dag&              45 \\\hline
Connect-4       &        84,124&      \l{94,077}&           13,664& Connect-4             &     \r{190,171}\ns&       63,427\nda&          45,894 \\\hline
English draughts&    \r{14,262}&           8,135&              872& English draughts-split&      \r{23,361}*  &        7,111\dag&           3,466 \\\hline
Gomoku          &         2,212&      \l{42,985}&              927& Gomoku-free style     &           2,430*  &   \l{26,878}\nda&      \emph{n/a} \\\hline
Hex             &         5,787&      \l{11,077}&       \emph{n/a}& Hex                   &           6,794\ns&   \l{10,625}\nda&      \emph{n/a} \\\hline
Reversi         &         2,012&       \l{2,081}&             203 & Reversi               &       \r{8,682}\ns&        1,312\nda&             373 \\\hline
The mill game   &         7,423&      \l{72,734}&       \emph{n/a}& The mill game         &      \r{10,102}*  &        2,467\dag&      \emph{n/a} \\\hline
Tic-tac-toe     &       400,000&     \l{535,294}&           85,319& Tic-tac-toe           &     \r{526,930}\ns&      422,836\nda&         104,500 \\\hline
\end{tabular}\end{center}
\begin{flushright}
*\ This game code was not originally available in RBG 1.0 and was added later.\\
\dag\ The rules in Ludii differ from the others.
\end{flushright}
\end{table*}

Based on our analysis of the benchmark in~\cite{piette2019empirical}, we have discovered several schemes that could occur when attempting cross-language comparison.
Here, we mention three most important of our findings.

\begin{enumerate}
\item First, what influenced the results the most is that the majority of the compared games do not have the same rules in all the three GGP systems. 
As we do not know the rules used in the previous Ludii's version, we base on the knowledge that correct rules of the concerned games in a proper variant, fully corresponding to those in RBG~1.0, were not present in public Ludii versions that soon followed the paper, and in most cases they are still not available.
The differences in the other games were usually based on simplifying the computation in favor of Ludii.

\item The repeated experiment for the RBG part, using exactly the same code, in most cases gave similar results, provided the hardware differences.
However, there are two exceptions, i.e., connect-4 and reversi, where the reported results were respectively about 2 and 4 times smaller than we could expect in our benchmark.
\item The results for GDL were obtained on different hardware than those for RBG and Ludii.
Indeed, exactly the same GDL results were reported before in~\cite{piette2019ludii}, where they were obtained with a slower processor, and the performance differences are also visible in our results.
Although the hardware used for GDL was not directly specified, we definitely see putting these results in the same table with the others to be misleading.
Furthermore, the result for chess was produced using the GGP-Prover instead of a propnet, despite that there was available an efficient chess implementation working well under a propnet.
\end{enumerate}

We note that some of the above-mentioned problems have been fixed in a more recent  update\footnote{\url{http://ludeme.eu/outputs/AAAI_20-6.pdf}}, with better matching of games and more accurate computation of RBG.
Sadly, some of the issues pointed out in~\cite{Kowalski2019ANote} still apply. 
Nevertheless, indeed, the conclusions drawn from the newer experiment are in the opposite to those from the older work and mostly agree with those from our tests.

Finally, we note that in~\cite{piette2019empirical}, there is also a comparison of game descriptions between Ludii, RBG, and GDL, which aims to estimate the clarity and simplicity of those languages.
The number of tokens (as a measure of description complexity) was calculated for each language.
However, the method of calculating is unclear enough, so that we were not able to find out the algorithm.
It has not been stated anywhere in the text, and all straightforward approaches that we tried to reproduce the method resulted in different numbers.

\subsection{Findings}

As for this branch of GGP research there is even less common ground between various approaches, there is also harder to find solid connections between each paper that will somehow force maintaining some standards.

Early works can be characterized by extremely small experimental sections, providing only a few results, usually not documented in detail. The trivial reason, being somewhat a justification, is that it is the consequence of the other authors' papers on which they based on. When a GGP language is introduced providing only 3 example of games with no more games to be found at any available repository, any approach to work with that language requires either to codify new games in it or to stick with the existing ones only.
This also leads to the next question -- is the primary work descriptive enough so that we can define our own games? What if it does not provide gamechecker, etc.? Sadly, such situations happen, but fortunately, there is definitely less of them as time goes by.

When comparing game-playing algorithms or the efficiency of search engines, obviously, information about the system specifications and the algorithm used are crucial.
But the mentioned research reveals a more subtle cause that can have a tremendous impact on the final results -- a proper game matching.
What is usually not so important from the human point of view when just thinking about the game may be very important from the algorithmic perspective. For many well-known games, it is surprisingly hard to specify what the ,,standard'' rules actually are.
Under the same name, often many rule variations can be hidden.
Obviously, the compared games should be the same, while what does it mean may be already not so obvious.
We think that the most natural definition of ``being the same'' could be ``have isomorphic game trees'', because then there exists a straightforward translation between playing these games.
However, specifying the rules or referring to exact implementations is still a rare practice.
Most of the research use common names without specifying the details, or even change the underlying rules of the game from one publication to another, yet still refer to it as it would have the same rules.

Finally, a commonsense assumption is that for both languages the game is encoded in an optimal way. Thus, when multiple game versions are available in the repository, the best-performing one should be used, and furthermore, using the best-known algorithm.

\section{Conclusion}

Based on the findings, we provide the following recommendations with explanations.
We hope that they will positively affect communication and cooperation in the community, and help to produce high-quality research on which further developments can be built collaboratively and safely.

Of course, apart from our domain-based conclusions, all the standard good research habits apply, as making source code openly available and well documented, or sharing easy-to-run test package to allow independent result reproduction, including more detailed recommendations that can be found in \cite{gundersen2018reproducible}.

\emph{(1) Provide the exact game descriptions that were used.}
Particularly in the cases when games in GDL were compared, usually only the name of the game was provided, despite the existence of various substantially different implementations.

\emph{(2) When comparing different game descriptions of the same game, make sure that they are equivalent.}
In the best case, provide a definition (e.g., by the same games, we understand those with isomorphic game trees).
Testing whether a game description is correct appears to be a difficult task.
Errors often occur in all GGP systems, sometimes left unnoticed for years.\footnote{Recently, we have found a subtle bug in all existing GDL implementations of the amazons game, commonly used in many experiments. A fixed version was proposed.}
To test the correctness, we propose commonly performing two kinds of automatic tests: \textbf{perft}, computing the number of all legal playouts up to a fixed depth, and \textbf{flat MC}, which provides statistics like the average goals, depth, and moves.
Furthermore, for popular games, the values could be published to allow verification of the correctness of new implementations (e.g., the perft results for chess are available at oeis\footnote{\url{https://oeis.org/A048987}}).

\emph{(3) Choose the testset that maximally covers the ones used in correlated research.}
Given limited time and resources, the authors usually prioritize examples for which they have the best results or the results of which give the best overview of their method. 
However, we advise to also take into account testcases that were considered in earlier works, as it sets the results more in context and helps followers to better estimate their own work compared to what is already existing.

\emph{(4) Provide the system specification and key software.}
This applies to the operating system and compiler or JVM versions.

\emph{(5) Run the test on a maximally idle system.}
It depends on the particular test. It is well known that processes in the background can influence the efficiency of a computation. Hence, to make the results reliable and maximally repetitive, it is better to avoid any background work.

\emph{(6) If the test should run on only one core, force this additionally to avoid misuse.}
In some cases, a single-threaded program does not really imply that only one core is used. This is common in modern Java, which, e.g., runs garbage collection on separate threads.
Since the real load of the system is then larger, it can cause unfair comparison with native programs that truly are executed on one core.
On linux systems, enforcing a one-core run can be achieved through \texttt{taskset} command.
Surprisingly, this can cause Java programs running faster and consuming less memory, as well as slower, depending on the test.
In any case, a note of whether the one-core restriction was used should be given, to avoid miscomparison when reproduced.

\emph{(7) If possible, when comparing with an existing implementation, ask the authors to ensure the correct usage.}
Although this may sound childish, it is still applicable  as concerns surprisingly many cases -- from being unsure of an answer received, up to the desire of not revealing own plans and intentions regarding other's work.
Irrespective of the reason, avoiding such contact reduces the quality of the conducted research.
Many systems are not documented well enough to ensure its proper usage, guaranteeing, e.g., their maximum performance.

\section*{Acknowledgments}

We thank Chiara F.\ Sironi for sharing the GDL propnet code and for helping with using it.
This work was supported by the National Science Centre, Poland under project number 2017/25/B/ST6/01920.

\bibliographystyle{aaai}
\bibliography{RAI2020}

\end{document}